%% file: root.tex
\pdfoutput=1

\RequirePackage{amsmath}         

\documentclass[letterpaper, 10 pt, conference]{ieeeconf}  
\IEEEoverridecommandlockouts                              

\makeatletter
\let\NAT@parse\undefined
\makeatother                               

\let\labelindent\relax                                    

\usepackage[per=frac,binary-units=true]{siunitx}
\usepackage{graphicx}            
\usepackage[caption=false, font=footnotesize]{subfig}
\usepackage{bm}
\usepackage{amssymb}             
\usepackage{amsfonts}            
\usepackage{enumitem}            
\usepackage{multirow}            
\usepackage{url}                 
\usepackage{xspace}              
\usepackage[T1]{fontenc}         
\usepackage{units}
\usepackage[utf8]{inputenc}
\usepackage{cleveref}
\usepackage{balance}
\usepackage{hyperref}

\usepackage[pdftex,dvipsnames]{xcolor}

\usepackage{tikz}
\usetikzlibrary{arrows}
\usetikzlibrary{positioning,calc}
\usetikzlibrary{decorations.pathreplacing}
\usetikzlibrary{decorations.markings}
\usetikzlibrary{fit}
\usetikzlibrary{shapes.callouts}
\usetikzlibrary{shapes.geometric}
\usetikzlibrary{matrix}

\usepackage[comma,numbers,compress]{natbib}

\graphicspath{{images/}}
\DeclareGraphicsExtensions{.pdf,.png,.jpg,.jpeg}

\title{\LARGE \textbf{Autonomous Dual-Arm Manipulation of Familiar Objects}}

\author{Dmytro Pavlichenko, Diego Rodriguez, Max Schwarz, Christian Lenz, Arul Selvam Periyasamy \\and Sven Behnke
\thanks{All authors are with the Autonomous Intelligent Systems (AIS) Group, Computer Science Institute VI,
        University of Bonn, Germany. Email: {\tt\small pavlichenko@ais.uni-bonn.de}.
        This work was supported by the European Union’s Horizon 2020 Programme under Grant Agreement 644839 (CENTAURO) and the German Research Foundation (DFG) under the grant BE 2556/12 ALROMA in priority programme SPP 1527 Autonomous Learning.}}

\usepackage{eso-pic}

\AtBeginDocument{\AddToShipoutPictureFG*{\AtTextUpperLeft{\put(0,\LenToUnit{9pt}){\parbox{\textwidth}{\centering\bfseries
International Conference on Humanoid Robots (Humanoids), Beijing, China, 2018
}}}}}
\begin{document}
	

\maketitle
\thispagestyle{empty}
\pagestyle{empty}

\begin{abstract}
	Autonomous dual-arm manipulation is an essential skill to deploy robots in unstructured scenarios.
	However, this is a challenging undertaking, particularly in terms of perception and planning.
	Unstructured scenarios are full of objects with different shapes and appearances that have to be grasped in a very specific manner so they can be functionally used.
	In this paper we present an integrated approach to perform dual-arm pick tasks autonomously.
	Our method consists of semantic segmentation, object pose estimation, deformable model registration, grasp planning and arm trajectory optimization.
	The entire pipeline can be executed on-board and is suitable for on-line grasping scenarios.
	For this, our approach makes use of accumulated knowledge expressed as convolutional neural network models and low-dimensional latent shape spaces.
	For manipulating objects, we propose a stochastic trajectory optimization that includes a kinematic chain closure constraint.
	Evaluation in simulation and on the real robot corroborates the feasibility and applicability of the proposed methods on a task of picking up unknown watering cans and drills using both arms.
\end{abstract}

\input{introduction.tex}

\input{related_work.tex}
\input{system_overview.tex}
\input{perception.tex}

\input{manipulation_planning.tex}
\input{evaluation.tex}

\input{conclusions.tex}

\bibliographystyle{IEEEtranN}
\bibliography{bibliography}

\end{document}

%% file: introduction.tex
\section{Introduction}

Daily-life scenarios are full of objects optimized to fit anthropometric sizes.
Thus, human-like robots are the natural solution to be used in quotidian environments.
In these scenarios, many objects require two or more grasping affordances in order to be manipulated properly.
Such objects may have complex shapes involving multiple degrees of freedom (DOF), be partially or completely flexible or simply be too large and/or heavy for single-handed manipulation,
for instance, moving a table and operating a heavy power drill.

In this paper, we describe an integrated system capable of performing autonomous dual-arm pick tasks.
Such tasks involve the consecutive accomplishment of several sub-tasks: object recognition and segmentation, pose estimation, grasp generation, and arm trajectory planning and optimization.
Each of these subproblems is challenging in unstructured environments when performed autonomously---due to the high level of uncertainty coming from noisy or missing sensory measurements, complexity of the environment, and modeling imperfection.
Thus, designing and combining software components which solve these sub-problems into one integrated pipeline is challenging.

We use semantic segmentation to detect the object.
A segmented point cloud is then passed to the next step of the pipeline: deformable model registration and grasp generation.
Since instances of the same object category are similar in their usage and geometry, 
we transfer grasping skills to novel instances based on the typical variations of their shape.
Intra-classes shape variations are accumulated in a learned low-dimensional latent shape space and are used to infer new grasping poses.

Finally, we optimize the resulting trajectories of the grasp planner by applying a modified version of Stochastic Trajectory Optimization for Motion Planning (STOMP) \cite{Kalakrishnan2011},
which we refer to as STOMP-New~\cite{Pavlichenko2017}.
We extend our previous work by adding an additional cost component to preserve the kinematic chain closure constraint when both hands hold an object.
For typical human-like upper-body robots, the dual-arm trajectory optimization problem with closure constraint is a non-trivial task due to curse of dimensionality and severe workspace constraints for joint valid configurations.
We perform experiments to investigate the influence of the new constraint on the performance of the algorithm.

\begin{figure}[t!]
    \centering
        \includegraphics[width=.9\linewidth]{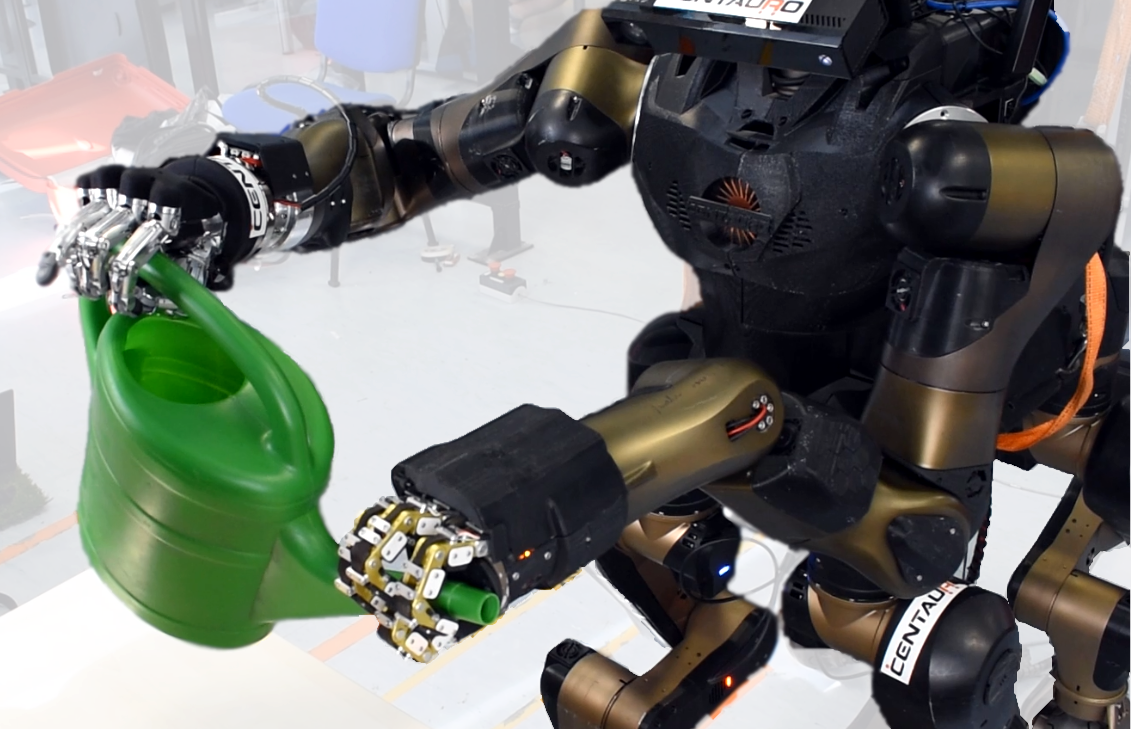}
        \caption
        {The Centauro robot performing bimanual grasping of a novel watering can.}
        \label{fig:teaser}
\end{figure}

The main contribution of this paper is the introduction of a complete software pipeline capable of performing autonomous dual-arm manipulation.
The pipeline was demonstrated with the Centauro robot~\cite{klamt2018supervised}.
Even though the robot base is quadruped, the upper-body is anthropomorphic with a torso, two arms, and a head.
We evaluate the capabilities of the designed system on the dual-arm pick task in simulation and on the real robot (Fig.~\ref{fig:teaser}).

%% file: related_work.tex
\section{Related Work}

Robotic systems which perform dual-arm manipulation are widely used for complex manipulation tasks.
Many of such systems are applied in industrial scenarios.
For instance, \citet{Kruger2011} present a dual arm robot for an assembly cell.
The robot is capable of performing assembly tasks both in isolation and in cooperation with human workers in a fenceless setup.
The authors use a combination of online and offline methods to perform the tasks.
Similarly, \citet{Tsarouchi2014} allow dual arm robots to perform tasks, which are usually done manually by human operators in a automotive assembly plant.
\citet{Stria2014} describe a system for autonomous real-time garment folding.
The authors introduce a new polygonal garment model, which is shown to be applicable to various classes of garment.
However, none of the previously mentioned works present a complete and generic pipeline, \citep{Kruger2011} and \citep{Tsarouchi2014} do \citet{Stria2014} was proposed a very specific and limited use-case.
To the best knowledge of the authors, there are no significant recent works, which present a complete autonomous robotic system for dual-arm manipulation.
In the following subsections we briefly review some of the noticeable works for each of the core components of our pipeline.

\subsection{Semantic Segmentation}
The field of semantic segmentation experienced much progress in recent years due to the availability of  large datasets.
Several works showed good performance using complex models that require extensive training on large data sets \cite{lin2016refinenet, Chen2018}.
In contrast, in this work we use a transfer learning method that focuses on fast training, which greatly increases the flexibility of the whole system~\cite{schwarz2018fast}.

\subsection{Transferring Grasping Skills}
\citet{Vahrenkamp2016} transfer grasp poses from a set of pre-defined grasps based on the RGB-D segmentation of an object.
The authors introduced a transferability measure which determines an expected success rate of the grasp transfer.
It was shown that there is a correlation between this measure and the actual grasp success rate.
In contrast, \citet{Stouraitis2015} and \citet{Hillenbrand2012} warp functional grasp poses such that the distance between point correspondences is minimized.
Subsequently the warped poses are replanned in order to increase the functionality of the grasp.
Those methods can be applied only in off-line scenarios, though, because of their large execution time.
The method explained here, on the other hand, is suitable for on-line scenarios.

\subsection{Dual-Arm Motion Planning}
Dual-arm motion planning is a challenging task, for which intensive research has been carried out.
\citet{Szynkiewicz2011} proposed an optimization-based approach to path planning for closed-chain robotic systems.
The path planning problem was formulated as a function minimization problem with equality and inequality constraints in terms of the joint variables.
\citet{Vahrenkamp2009} presented two different approaches for dual-arm planning: $J^+$ and IK-RRT.
Although the first one does not require an inverse kinematics (IK) solver, 
IK-RRT was shown to perform better on both single and dual-arm tasks.
In contrast, a heuristic-based approach was proposed by \citet{Cohen2014}.
The method relies on the construction of a manipulation lattice graph and an informative heuristic.
Even though the success of the search depends on the heuristic, the algorithm showed good performance in comparison with several sampling-based planners.
\citet{Byrne2015} proposes a method consisting of goal configuration sampling, subgoal selection and Artificial Potential Fields (APF) motion planning.
It was shown that the method improves APF performance for independent and cooperative dual-arm manipulation tasks.
An advantage of our approach to arm trajectory optimization is the flexibility of the prioritized cost function which can be extended to support different criteria, which we demonstrate in this work.

%% file: system_overview.tex
\section{System Overview}

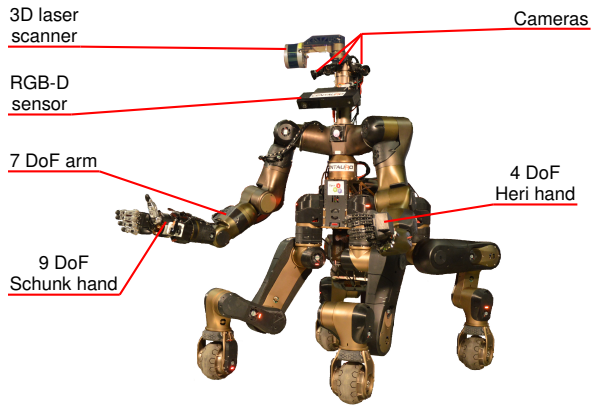
\begin{figure}
\centering
\scalebox{1.0}{\input{images/centauro_overview.pgf}}
\caption{The Centauro robot. The main components of the upper-body are labeled.}
\vspace{-0.5cm}
\label{fig:centauro_robot}
\end{figure}

In this work we test our software pipeline on a centraur-like robot, developed within the CENTAURO project\footnote{\url{https://www.centauro-project.eu/}}.
The robot has a human-like upperbody, which is mounted on the qudrupedal base.
It is equipped with two anthropomorphic manipulators with 7 DOF each.
The right arm possesses a SVH Schunk hand as an end-effector, 
while the left arm is equipped with a Heri hand~\cite{heri}.
The sensor head has a Velodyne Puck rotating laser scanner with spherical field of view as well as multiple cameras.
In addition, a Kinect v2~\cite{fankhauser2015} is mounted on the upper part of the chest.
The Centauro robot is depicted in Fig.~\ref{fig:centauro_robot}.

\begin{figure*}[ht!]
	\centering
	\includegraphics[width=\textwidth]{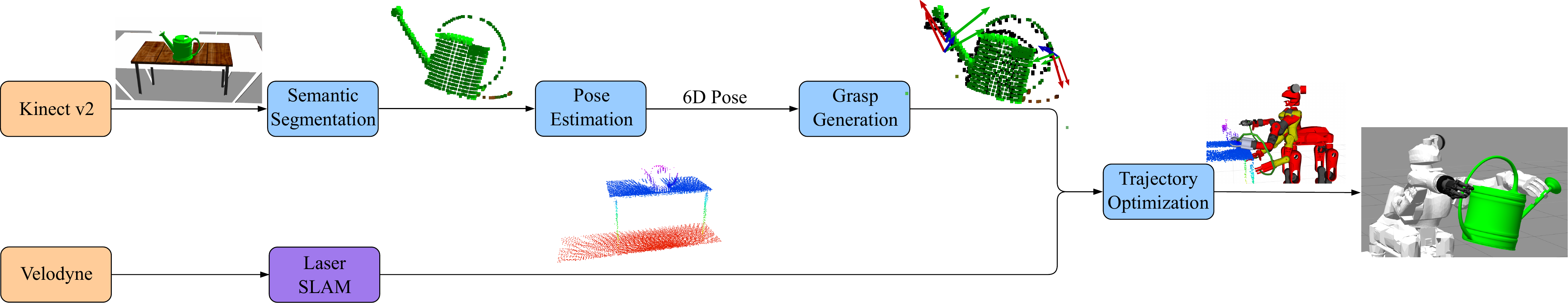}
	\caption
	{Simplified diagram of the system, showing the information flow between core components. Orange: sensors; Blue: main components of the pipeline: Purple: external modules.}
	\label{fig:system_overview}
\end{figure*}

In order to perform an autonomous dual-arm pick tasks we propose the following pipeline (Fig.~\ref{fig:system_overview}):
\begin{itemize}
 \item \textit{Semantic Segmentation} performed by using RGB-D data from the Kinect v2,
 \item \textit{Pose Estimation} on the resulting segmented point cloud,
 \item non-rigid \textit{Shape Registration} to obtain grasping poses,
 \item and finally, \textit{Trajectory Optimization} to obtain collision-free trajectories to reach pre-grasp poses.
\end{itemize}

%% file: images/centauro_overview.pgf
\begin{tikzpicture}[
 	font=\sffamily\footnotesize,
    every node/.append style={text depth=.2ex},
	box/.style={rectangle, inner sep=0.5, anchor=west},
	line/.style={red, thick},
	l/.style={font=\sffamily\scriptsize},
]


\node[anchor=south west,inner sep=0] (image) at (0,0) {\includegraphics[width=0.6\linewidth]{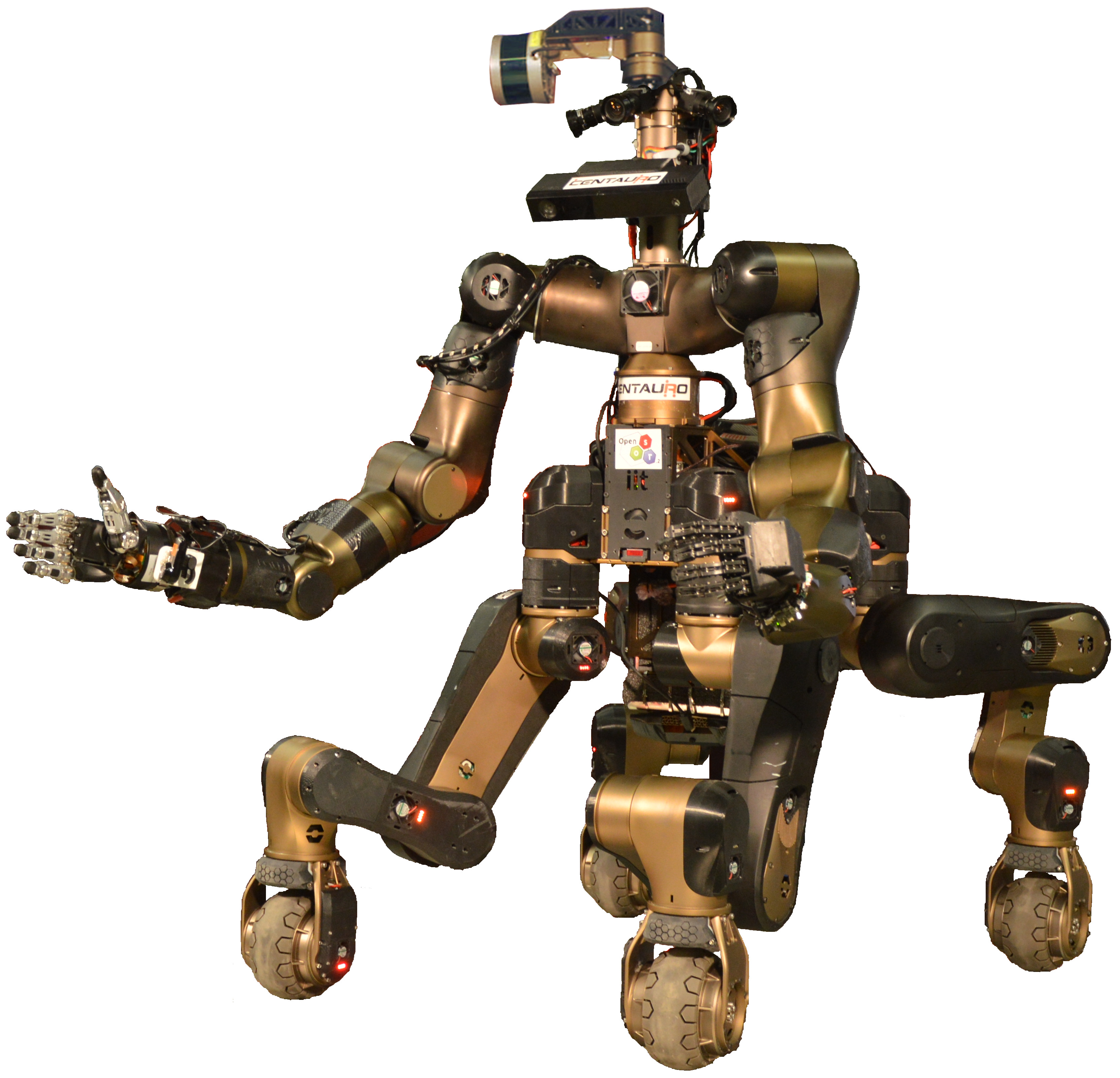}};

\node[box, align=center,l](laser_scanner) at(-1.4,5.1){3D laser\\scanner};
\draw[line](laser_scanner.south west)--(laser_scanner.south east);
\draw[line](laser_scanner.south east)--(2.3,4.8);

\node[box,l](cameras) at(5.3 ,5.2){Cameras};
\coordinate(camera_split) at (3.3,5);
\draw[line](cameras.south west)--(cameras.south east);
\draw[line](cameras.south west)--(camera_split);
\draw[line](camera_split)--(3.3, 4.6);
\draw[line](camera_split)--(2.7,4.5);
\draw[line](camera_split)--(3.0,4.6);

\node[box,align=center,l](kinect) at(-1.4,4.2){RGB-D\\sensor};
\draw[line](kinect.south west)--(kinect.south east);
\draw[line](kinect.south east)--(2.5,4.15);

\node[box,l](arm) at(-1.4,3.3){7 DoF arm};
\draw[line](arm.south west)--(arm.south east);
\draw[line](arm.south east)--(1.5,2.6);

\node[box,align=center,l](schunk) at(-1.4,1.8){9 DoF\\Schunk hand};
\draw[line](schunk.south west)--(schunk.south east);
\draw[line](schunk.south east)--(0.7,2.5);

\node[box,align=center,l](soft_hand) at(5.05,3){4 DoF\\Heri hand};
\draw[line](soft_hand.south west)--(soft_hand.south east);
\draw[line](soft_hand.south west)--(3.6,2.5);

%
%

\end{tikzpicture}
 

%% file: perception.tex
\section{Perception}
\label{sec:perception}

For perceiving the object to be manipulated, a state-of-the-art semantic segmentation architecture \citep[RefineNet]{lin2016refinenet} is trained on synthetic scenes.
Those are composed of a small number of captured background images which are augmented randomly with inserted objects.
This approach follows \citet{schwarz2018fast} closely, with the exception that the inserted object segments
are rendered from CAD meshes using the open-source Blender renderer.
The core of the model consists of four ResNet blocks.
After each block the features become more abstract, but also lose the resolution.
So, the feature maps are upsampled and merged with the map from the next level, until the end result is at the same time high-resolution and highly semantic feature map.
The final classification is done by a linear layer followed by a pixel-wise SoftMax.

At inference time, also following \citet{schwarz2018fast}, we postprocess the semantic
segmentation to find individual object contours. The dominant object is found using
the pixel count and is extracted from the input image for further processing.

The 6D pose of the object is estimated as follows: the translation component is computed by projecting the centroid of the object contour into 3D by using the depth information;
the orientation component is calculated from the principle components on the 3D object points of the object and incorporating prior knowledge of a canonical model defined for each category.
This initial pose estimate is refined by the shape space registration described in Sec. \ref{sec:grasping}.

%% file: manipulation_planning.tex
\section{Manipulation Planning}

\subsection{Grasp Planning}
\label{sec:grasping}
The grasp planning is a learning-based approach that exploits the fact that objects similar to each other can be grasped in a similar way.
We define a category as a set of models with related extrinsic geometries.
In the training phase of the method, a shape (latent) space of the category is built.
This is done by computing the deformation fields of a canonical model $\mathbf{C}$ towards the other models in the category.
This is carried out by using the Coherent Point Drift (CPD) non-rigid registration method.
CPD provides a dense deformation field, thus new points can be warped even after the registration.
Additionally, the deformation field of each object in the training set can be expressed in a vector whose dimensionality equals the number of points times the number of dimensions of the canonical model.
This mean that the variations in shape from one object to the other can be expressed by a vector of the same length across all training samples.
Thus, subspace methods can be straightforwardly applied. 
Finally, the principal components of all these deformation fields are calculated by using Principal Component Analysis - Expectation Maximization (PCA-EM) which define the basis of the shape space.

\begin{figure}[]
	\centering
	\includegraphics[width=\linewidth]{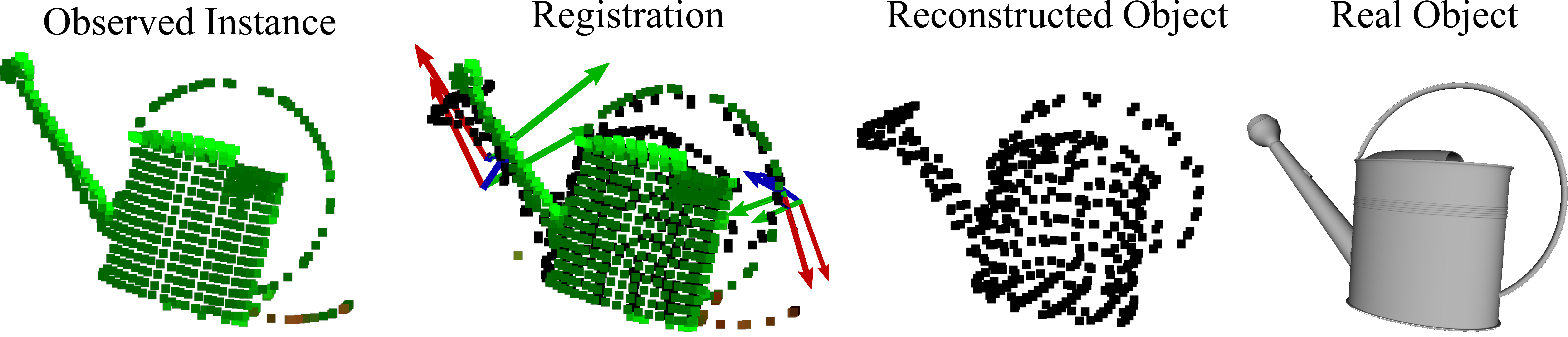}
	\caption
	{Shape space registration on the watering can category. The method is able to reconstruct a partially occluded instance containing noise.}
	\label{fig:registration}
\end{figure}

Once the shape space is constructed, new instances can be generated by interpolating and extrapolating in the subspace.
In the inference phase, we search in the latent space in a gradient-descent fashion for an instance which relates to the observed model at best.
We do this by optimizing a non-linear function that minimizes a weighted point distance.
An additional rigid registration is also incorporated in the cost function to account for misalignments.
Furthermore, the latent variables are regularized which has shown to provide numerical stability.
Once the descriptor in the latent space is known, it is transformed back to obtain the deformation field that best describes the observation.
In this process, partially occluded shapes are reconstructed.
The registration is robust against noise and misalignments to certain extent~\citep{Rodriguez2018a}.
Fig.~\ref{fig:registration} shows a partially observed instance with noise and the reconstructed object after the shape registration.

The canonical model has associated control poses that describe the grasping motion.
These control poses are warped using the inferred deformation field.
More details about the shape space registration can be found in \citep{Rodriguez2018b}.
For bimanual manipulation we associate individual grasping control frames to each arm and warp them according to the observed model.
Because each of the control poses is independent, simultaneous arm motions are possible.
The control poses contain the pre-grasp and final grasp poses.

\subsection{Trajectory Optimization}
The grasp planner provides pre-grasp poses for both arms, the trajectory optimizer plans a collision-free trajectory to reach them.
We use STOMP-New, which showed better performance in previous experiments~\cite{Pavlichenko2017}.
It has a cost function consisting of five cost components: collisions, joint limits, end-effector orientation constraints, joint torques and trajectory duration.
The input is an initial trajectory $\Theta$  which consists of $N$ keyframes ${\bm{\theta}_i \in \mathbb{R}^J}, i\in\{0,\ldots,N\num{-1}\}$ in joint space with $J$ joints.
Normally, a na{\"i}ve linear interpolation between the given start and goal configurations $\bm{\theta}_{start}$ and $\bm{\theta}_{goal}$ is used.
The start and goal configurations are not modified during the optimization.

Since the optimization is performed in joint space, extending the algorithm to use two arms instead of one is straightforward.
We extended the approach to support multiple end-effectors (two in the context of this work),
so trajectories of two independent arms are simultaneously optimized.

However, for moving an object grasped with two hands, a kinematic chain closure constraint has to be satisfied.
Thus, the following term $q_{cc}(.,.)$ is added to the cost function:
\begin{equation}\label{eq:cost_function}
\begin{aligned}
q(\bm{\theta}_{i}, \bm{\theta}_{i+1}) =  &q_{o}(\bm{\theta}_{i}, \bm{\theta}_{i+1}) + q_l(\bm{\theta}_{i}, \bm{\theta}_{i+1}) + q_{c}(\bm{\theta}_{i}, \bm{\theta}_{i+1}) \\ + &q_{d}(\bm{\theta}_{i}, \bm{\theta}_{i+1}) + q_{t}(\bm{\theta}_{i}, \bm{\theta}_{i+1}) + q_{cc}(\bm{\theta}_{i}, \bm{\theta}_{i+1}),
\end{aligned}
\end{equation}
where ${q(\bm{\theta}_i, \bm{\theta}_{i+1})}$ is a cost for the transition from the configuration $\bm{\theta}_i$ to $\bm{\theta}_{i+1}$.
The cost function now consists out of six terms, where the first five are coming from our original implementation of STOMP-New.
By summing up costs $q(\cdot,\cdot)$ of the consecutive pairs of transitions ${\bm{\theta}_{i}, \bm{\theta}_{i+1}}$ of the trajectory $\Theta$, we obtain the total cost.

The new term $q_{cc}(\cdot,\cdot)$ for the kinematic chain closure constraint is formulated as:
\begin{equation}
\resizebox{\linewidth}{!}
{ $q_{cc}(\bm{\theta}_{i}, \bm{\theta}_{i+1})=\frac{1}{2} \max\limits_{j} q_{ct}(\bm{\theta}_{j})  +  \frac{1}{2} \max\limits_{j} q_{co}(\bm{\theta}_{j}), j \in \{i,\dots,i+1\}$ }
\label{eq:closure_constraint}
\end{equation}
where $q_{ct}(\cdot)$ penalizes deviations in translation between the end-effectors along the transition and $q_{co}(\cdot)$ penalizes deviations of the relative orientation of the end-effectors.

Given two end-effectors, $eef_{1}$ and $eef_{2}$, the initial translation ${t_{desired} \in \mathbb{R}^3}$ between them is measured in the first configuration $\bm{\theta}_{0}$ of the trajectory.
Then, for each evaluated configuration $\bm{\theta}_{j}$, the corresponding translation $t_j$ between $eef_{1}$ and $eef_{2}$ is computed.
The deviation from the desired translation is thus defined as: ${\delta t = |t_{desired} - t_j|}$.
Finally, we select the largest component ${t_{dev} = \max\limits_{x,y,z} \delta t | \delta t = \langle x, y, z \rangle}$ and compute the translation cost:
\begin{equation}\label{eq:translation_cost}
\begin{aligned}
q_{ct}(\bm{\theta}_{j}) =
\begin{cases}
    C_{ct} + C_{ct} \cdot t_{dev} & \mbox{if } t_{dev} \geq t_{max} \\  
    \frac{t_{dev}}{t_{max}}, & \mbox{otherwise}
\end{cases},
\end{aligned}
\end{equation}
where $t_{max}$ is the maximum allowed deviation of the translation component and ${C_{ct} \gg 1}$ is a predefined constant.
Thus, ${q_{ct} \in [0,1]}$ if the deviation of the translation is below the allowed maximum and ${q_{ct} \gg 1}$ otherwise.

Similarly, we define the term $q_{co}(\cdot)$ for penalizing deviations in the orientation.
The initial relative orientation $o_{desired} \in \mathbb{R}^3$ between $eef_{1}$ and $eef_{2}$ is calculated in the first configuration $\bm{\theta}_{0}$.
For each configuration $\bm{\theta}_{j}$, the corresponding relative orientation $o_j$ is measured.
The deviation from the desired orientation is computed as: ${\delta o = |o_{desired} - o_j|}$.
We select the largest component ${o_{dev} = \max\limits_{r,p,y} \delta o | \delta o = \langle r,p,y \rangle}$ and compute the orientation cost:
\begin{equation}\label{eq:orientation_cost}
\begin{aligned}
q_{co}(\bm{\theta}_{j}) =
\begin{cases}
    C_{co} + C_{co} \cdot o_{dev} & \mbox{if } o_{dev} \geq o_{max} \\  
    \frac{o_{dev}}{o_{max}}, & \mbox{otherwise}
\end{cases},
\end{aligned}
\end{equation}
where $o_{max}$ is the maximum allowed deviation of the orientation component and $C_{co} \gg 1$ is a predefined constant.
Extending the algorithm with this constraint allows to optimize trajectories, maintaining the kinematic chain closure constraint,
and, hence, plan trajectories for moving objects which are held with two hands.

%% file: evaluation.tex
\section{Evaluation}
First, we present the evaluation of the arm trajectory optimization alone.
In the latter subsection, we evaluate the performance of the developed pipeline by picking a watering can with two hands in simulation.
Finally, we present the experiments performed with the real robot: dual-arm picking of watering can and drill.
\begin{figure}[]
	\centering
	\subfloat[]{\includegraphics[width=4.1cm]{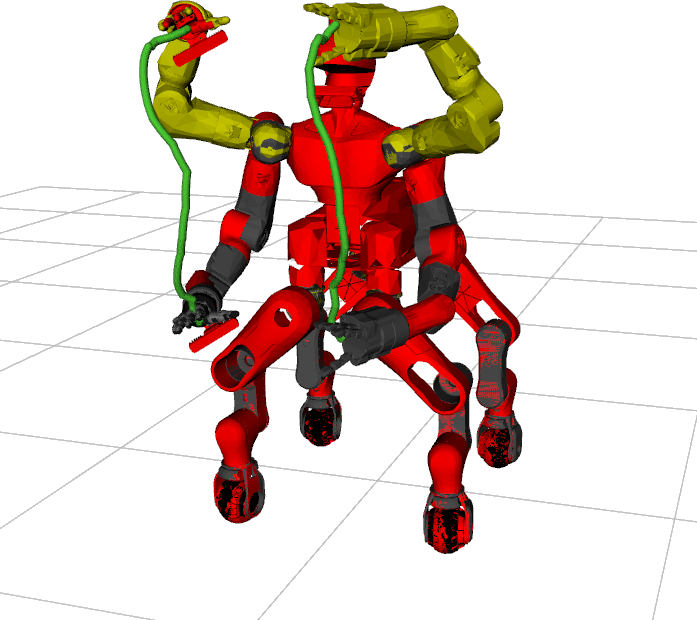}}\hfill
	\subfloat[]{\includegraphics[width=4.1cm]{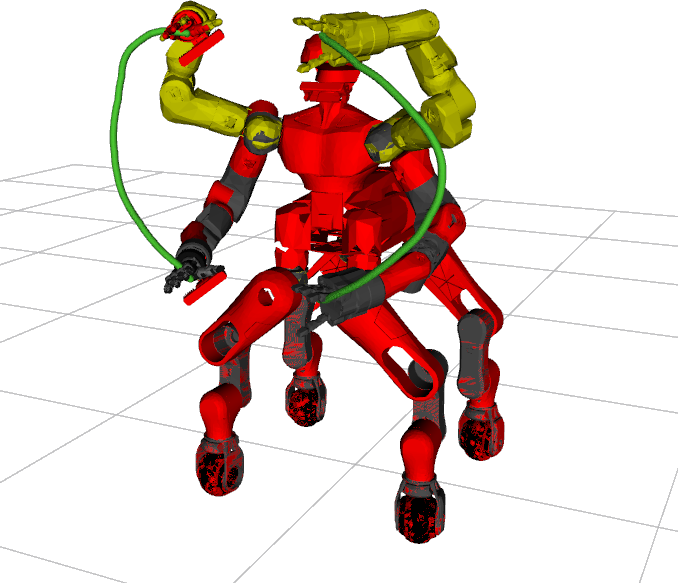}}
	\caption{Comparison of the trajectories obtained with/without kinematic chain closure constraint. Red: start configuration; Yellow: goal configuration; Green: paths of the end-effectors.
		\textbf{(a)} Closure constraint enabled. The robot has to follow the kinematically difficult path. \textbf{(b)} Closure constraint disabled. The arms can be moved easily to the sides of the robot.}
	\label{fig:traj_opt_comparison}
\end{figure}

\subsection{Trajectory Optimization}

\begin{figure*}
	\centering
	\includegraphics[width=4.cm]{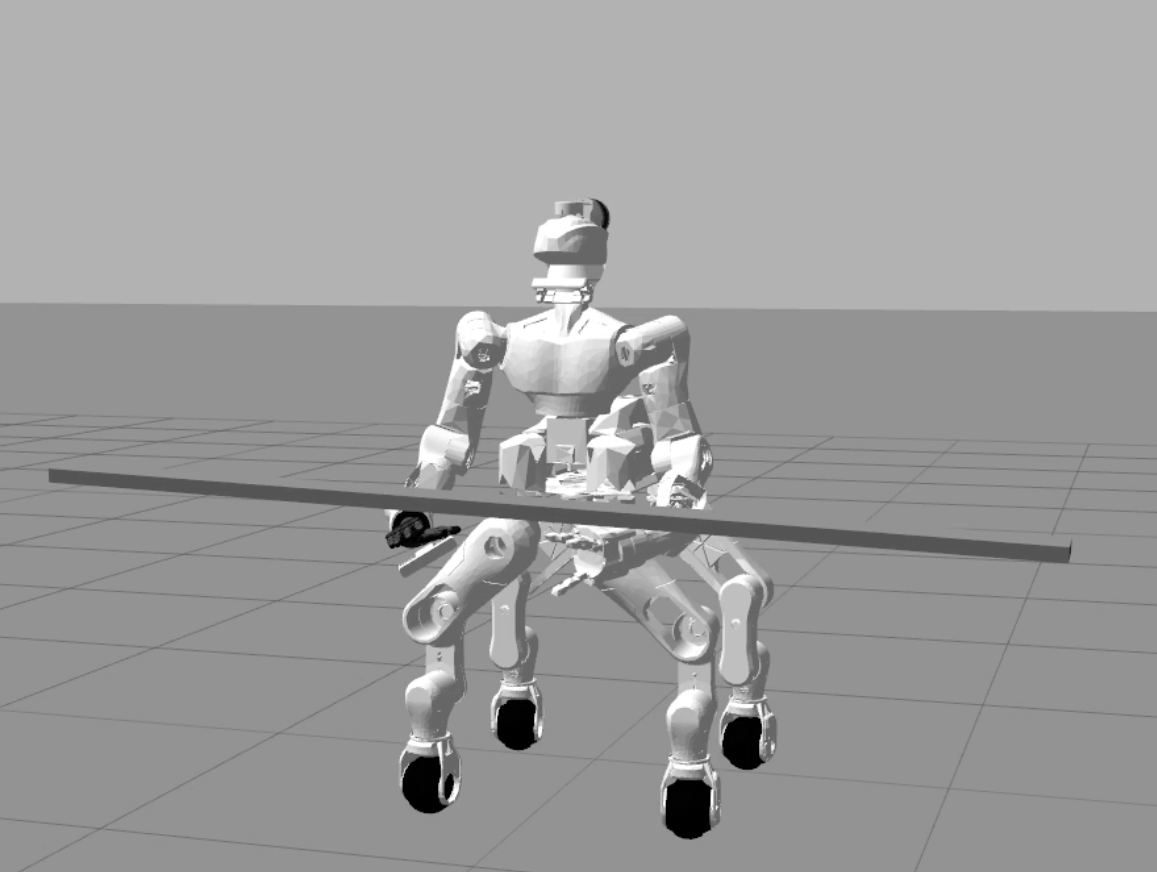}\hfill
	\includegraphics[width=4.cm]{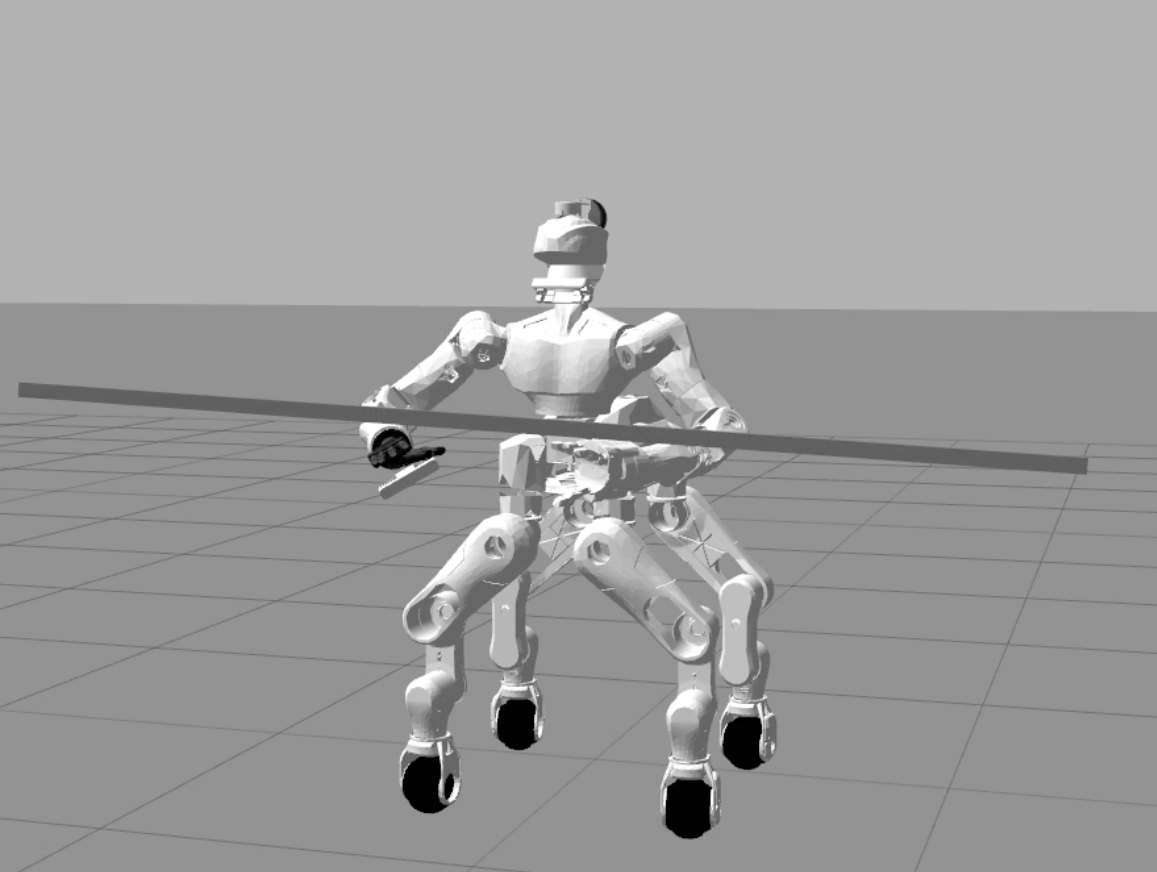}\hfill
	\includegraphics[width=4.cm]{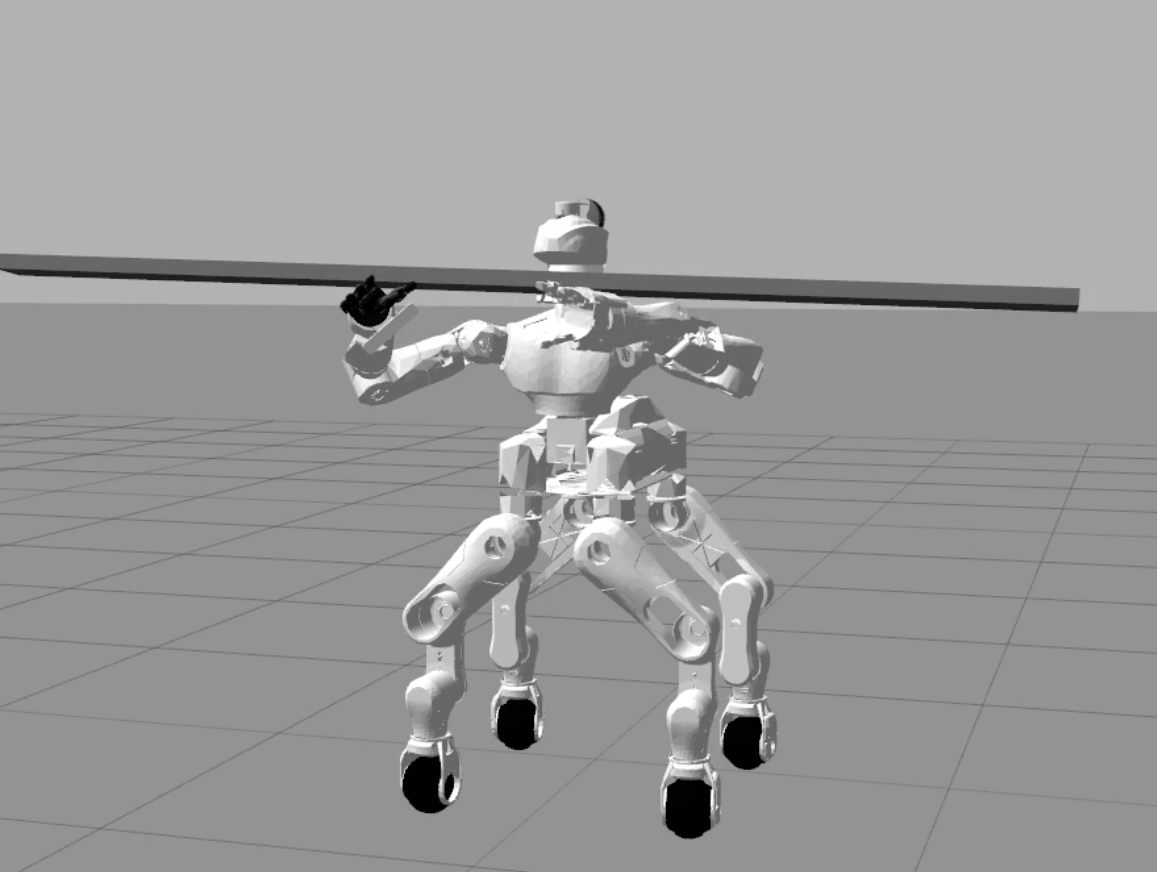}\hfill
	\includegraphics[width=4.cm]{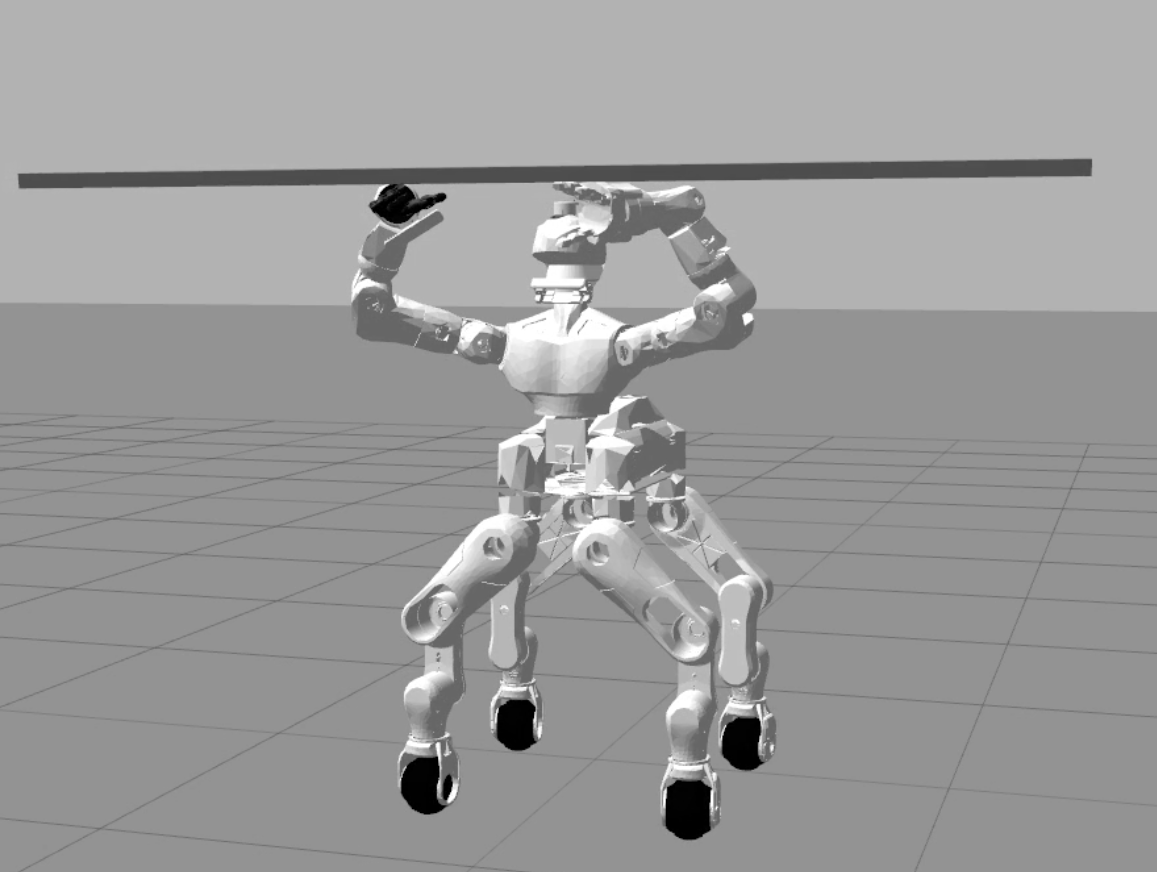}
	\caption{The Centauro robot lifting a long bulky bar. As the bar is laying on the wrists unsecured, not only the closure constraint has to be preserved, but also the orientation of the end-effectors
		has to remain the same during the whole trajectory.}
	\label{fig:lifting_bar}
\end{figure*}

Experiments were performed using the gazebo simulator with the Centauro robot.
Both 7\,DOF arms were used simultaneously, resulting in a total of 14\,DOF.
We performed the experiments on an Intel Core i7-6700HQ CPU, 16\,GB of RAM, 64\,bit Kubuntu 16.04 with 4.13.0-45 kernel using ROS Kinetic.
The algorithm ran on a single core with 2.60\,GHz.

We investigate how the introduction of the close chain kinematic constraint influences the performance of the algorithm.
We compared the performance of the algorithm with and without the constraint in an obstacle-free scenario, where the robot had to lift both arms upwards (Fig.~\ref{fig:traj_opt_comparison}).
We solved the problem 50 times with enabled/disabled closure constraint, each.
The time limit for the algorithm was set to 10\,s.
The obtained runtimes and success rates are shown in the Table~\ref{table:runtime}.

\begin{figure}[b!]
	\centering
	\includegraphics[width=0.9\linewidth]{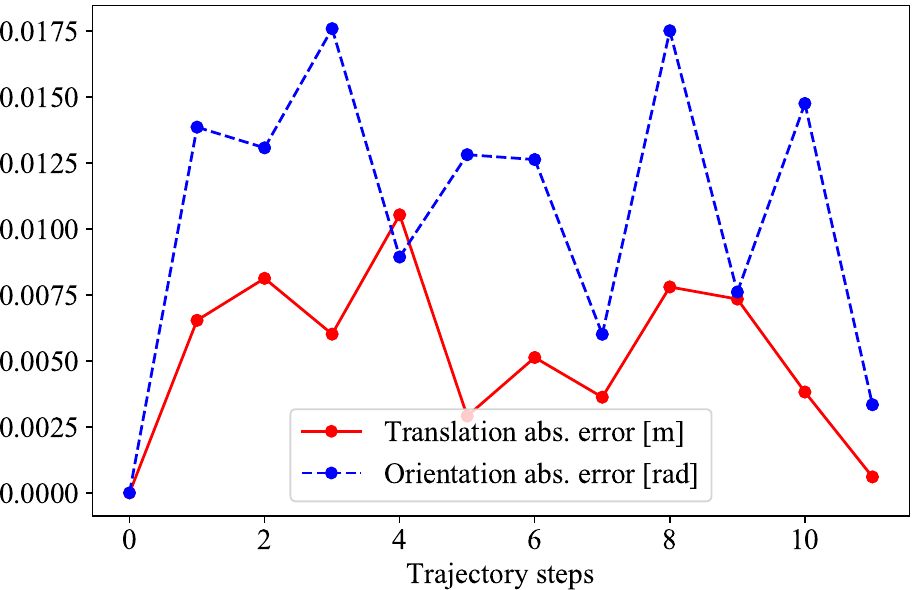}
	\caption
	{Error between desired and observed end-effectors relative pose for trajectories shown in Fig.~\ref{fig:traj_opt_comparison}.}
	\label{fig:constraint_error}
\end{figure}

\begin{table}[]
	\caption{Comparison of the average runtime and success rate with/without closure constraint.}
	\label{table:runtime}
	\begin{tabular}{ccc}
		\hline
		& \multicolumn{1}{l}{Without closure constraint} & \multicolumn{1}{l}{With closure constraint} \\ \hline
		Runtime {[}s{]}                    & 0.34$\pm$0.01                                  & 4.31$\pm$2.42                               \\ \hline
		Success rate                       & 100\%                                          & 83\%                                        \\ \hline
		\multicolumn{1}{l}{Runtime growth} & ---                                            & 1267\%                                      \\ \hline
	\end{tabular}
\end{table}
When the algorithm performs optimization without closure constraint, the runtime is relatively short with a very small standard deviation and 100\% success rate.
On the other hand, with enabled closure constraint, the runtime grew significantly by 1267\% and the success rate dropped to 83\%.
This happens because the space of valid configurations is largely reduced when enforcing the closure constraint and the sampling-based algorithm struggles to converge to a valid solution.
This also explains the large standard deviation for the case when the closure constraint is enabled.
In Fig.~\ref{fig:constraint_error} the error between desired and actual pose of the end-effectors, observed during one of those trajectories, is shown.

We also demonstrate the optimization with closure constraints enabled for a practical task.
The robot has a long bulky bar laying on its wrists (Fig.~\ref{fig:lifting_bar} (a)) and the task is to lift it up.
Since the bar is not secured in any way, it is not only necessary to preserve the closure constraint, but also to maintain the exact orientation of the end-effectors along the whole trajectory (Fig.~\ref{fig:lifting_bar}).

\subsection{Dual-Arm Picking in Simulation}
We evaluate the proposed system by picking a watering can with two arms in a functional way, 
i.e., that the robot can afterwards use it.
The experiments were performed in the Gazebo simulator with the Centauro robot.
To speed up the simulation, only the upper-body was actuated.
Moreover, the collision model of the fingers were modeled as primitive geometries: capsules and boxes.
The laser scanner and the RGBD sensor were also incorporated in the simulation.
We trained the semantic segmentation model using synthetic data. 
We used 8 CAD models of the watering can to render 400 frames.
Additional training data with semantic labeling is obtained by placing the frames onto multiple backgrounds and generating the ground truth labels. 

\begin{table}[]
	\centering
	\caption{Success rate of picking watering cans from the test set and performance of the trajectory optimization method.}
	\label{table:traj_opt_comparison}
	\begin{tabular}{cc|c}
		\hline
		& \begin{tabular}[c]{@{}c@{}}Success rate\\ (attempts to solve)\end{tabular} & \begin{tabular}[c]{@{}c@{}}Traj. opt. runtime {[}s{]}\\ Success rate\end{tabular} \\ \hline
		Can 1 & 75\% (4)                                                                   & \multirow{3}{*}{\begin{tabular}[c]{@{}c@{}}0.9$\pm$0.24\\ 100\%\end{tabular}}     \\ \cline{1-2}
		Can 2 & 100\% (5)                                                                  &                                                                                   \\ \cline{1-2}
		Can 3 & 60\% (3)                                                                   &                                                                                   \\ \hline
	\end{tabular}
\end{table}
\begin{figure}[b!]
	\centering
	\includegraphics[width=5.5cm]{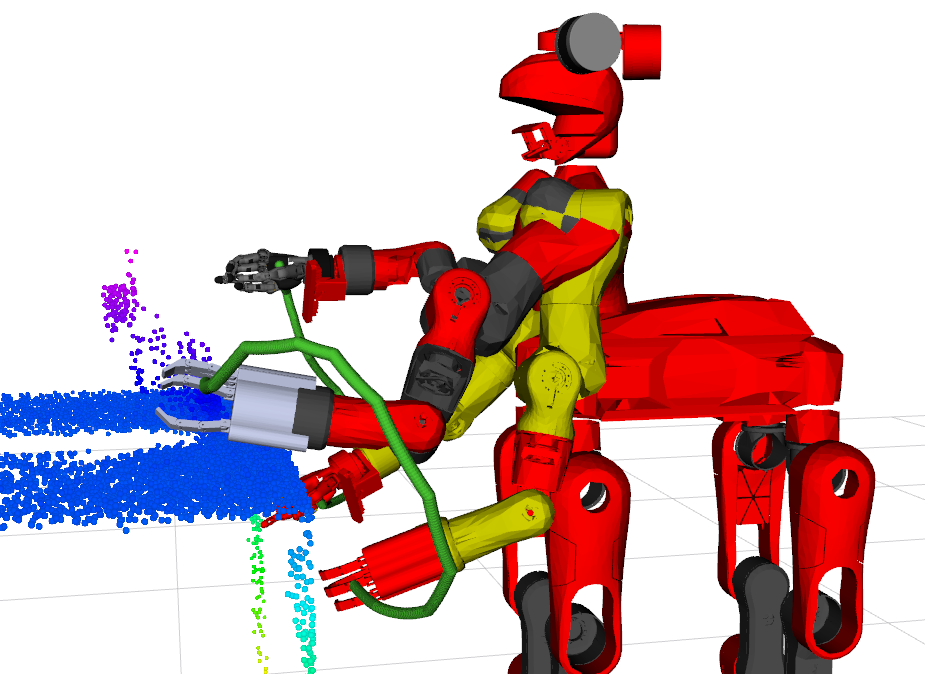}
	\caption
	{Dual-arm trajectory for reaching pre-grasp poses. Yellow: initial pose; Black and grey: goal pose; Green: paths of the end-effectors.
		The arms have to retract back in order to avoid collisions with the table.}
	\label{fig:trajectory_example}
\end{figure}

\begin{figure*}[]
	\centering
	\subfloat[]{\includegraphics[width=3.4cm]{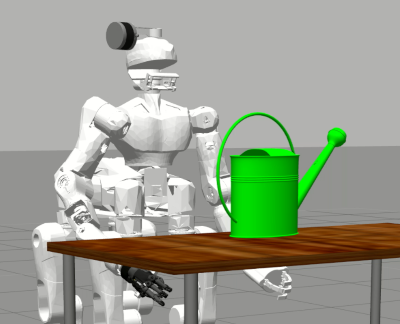}}\hfill
	\subfloat[]{\includegraphics[width=3.4cm]{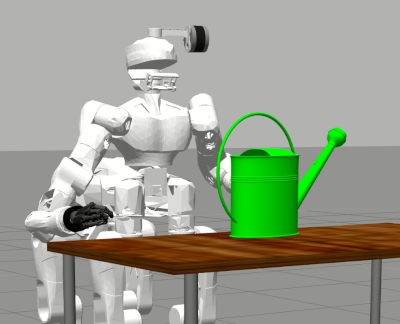}}\hfill
	\subfloat[]{\includegraphics[width=3.4cm]{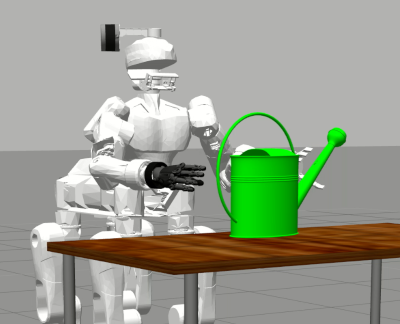}}\hfill
	\subfloat[]{\includegraphics[width=3.4cm]{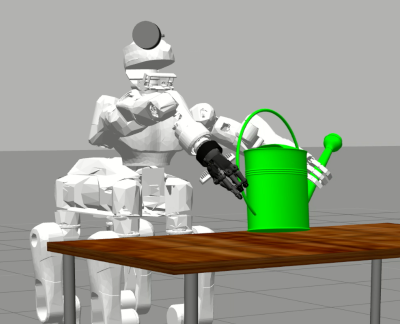}}\hfill
	\subfloat[]{\includegraphics[width=3.4cm]{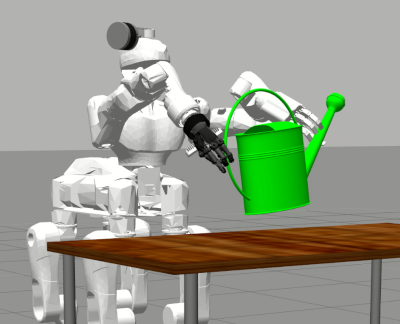}}
	\caption{Centauro performing a dual-arm functional grasp of the watering can in simulation. \textbf{(a)} Initial pose. \textbf{(b)} - \textbf{(c)} Reaching the pre-grasp pose. 
		\textbf{(d)} Can is grasped. \textbf{(e)} Can is lifted.}
	\label{fig:can_lifting}
\end{figure*}

For constructing the shape space we define a training set composed of the same watering cans used to train the semantic segmentation model.
The test set consisted out of three different watering cans.
For the registration, the objects were represented as point clouds generated by ray-casting operations on meshes obtained from 3D databases.
The shape space contained 8 principal components.

The task of the experiment is to grasp and to lift upwards all three cans from the test set.
Each trial starts with the robot standing in front of the table, on which the watering can is placed.
The arms of the robot are located below the surface of the table, 
so that a direct approach (straight line) to the object will result in a collision.
Each can had to be successfully grasped three times with different orientation so that the task is considered solved.
In this manner, the can is rotated around its Z-axis for +0.25, 0 and -0.25 radians.
In order to evaluate the performance of the non-rigid registration against misalignments,
noise in range $\pm 0.2$ radians was added to the yaw component of the 6D pose.
The trials were performed until each of the three grasps succeeded once.
Obtained success rates and measured average runtime of the trajectory optimization method are presented in Table~\ref{table:traj_opt_comparison}.

\begin{figure}[]
	\centering
	\subfloat[]{\includegraphics[width=2.7cm]{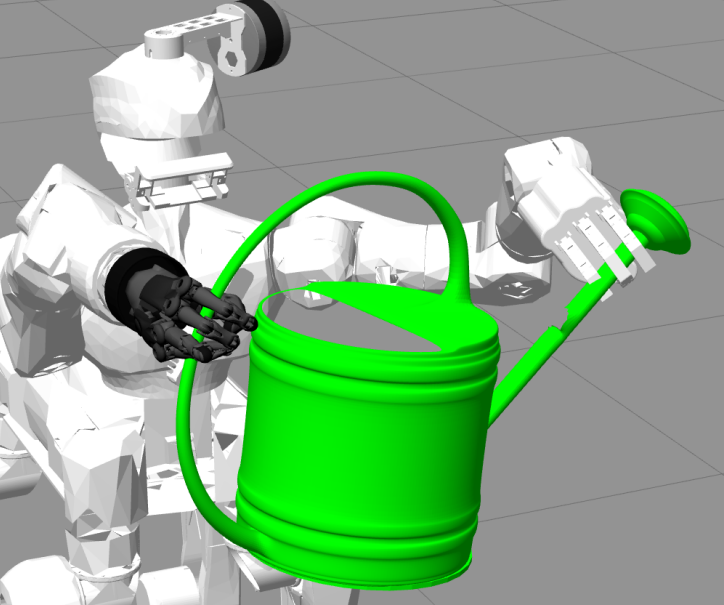}}\hfill
	\subfloat[]{\includegraphics[width=2.7cm]{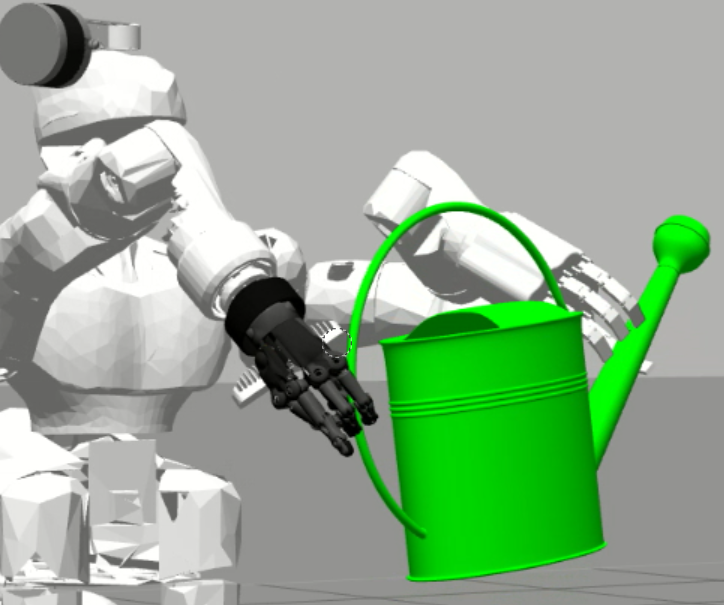}}\hfill
	\subfloat[]{\includegraphics[width=2.7cm]{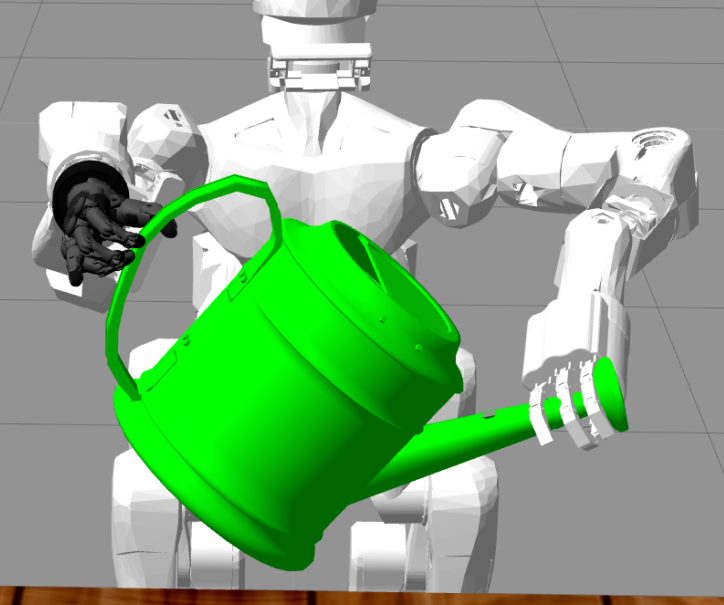}}
	\caption{Three cans from the test set successfully grasped.\textbf{(a)} - \textbf{(c)} Can 1, Can 2, Can 3, respectively. Note that all the cans have different geometry.}
	\label{fig:three_cans_grasped}
\end{figure}

Our system solved the task Can 2 with no issues,
whereas Can 1 and especially Can 3 were more difficult.
For Can 1, there was a minor misalignment of the grasp pose for the right hand, which did not allow us to grasp the can successfully.
Can 3 had the most distinctive appearance among the cans in our dataset, that is why it caused the most difficulties.
During the experiment we often had to run the non-rigid registration several times because it was stuck in local minima.
STOMP-New showed consistent success rate and satisfactory runtime of around one second.
Typical trajectories for reaching pre-grasp poses are shown in Fig.~\ref{fig:trajectory_example}.
The Centauro robot performing the experiment with Can 2 is depicted in Fig.~\ref{fig:can_lifting}.
All three cans forming our test set, successfully grasped, are shown in Fig~\ref{fig:three_cans_grasped}.

\subsection{Real-Robot Experiments}
On the real Centauro robot we performed the same experiment, as described above for a single orientation of the watering can.
The pipeline was executed five times in attempt to grasp the can with two hands in a functional way.
The method succeeded four times out of five.
We measured the average runtime for each component of the system as well as the success rate (Table~\ref{table:whole_pipeline}).

\begin{table}[]
	\centering
	\caption{Average runtime and success rate of each component of the pipeline.}
	\begin{tabular}{ccc}
		\hline
		Component               & Runtime {[}s{]} & Success rate \\ \hline
		Semantic segmentation   & 0.74       & 100\%        \\ \hline
		Pose estimation         & 0.12       & ---          \\ \hline
		Grasp generation        & 4.51 $\pm$ 0.69 & ---          \\ \hline
		Trajectory optimization & 0.96 $\pm$ 0.29 & 100\%        \\ \hline
		Complete pipeline       & 6.27 $\pm$ 0.98 & 80\%         \\ \hline
	\end{tabular}
	\label{table:whole_pipeline}
\end{table}

We do not provide the success rate for the pose estimation, since the ground truth was not available.
Consequetly, it is hard to assess the success rate of grasp generation as it may fail due to the previous step of the pipeline.
Trajectory optimization method shown a consistent average runtime of around 1\,s and a 100\% success rate.
Overall, the pipeline took around 6\,s on average with a success rate of 80\%.
One of the attempts failed on the stage of grasping the can,
because the approaching (goal) pose of the trajectory optimizer was not close enough to the object which resulted in a collision between the hand and the watering can while reaching the pregrasp pose. 
Consequently, the object moved away from the estimated pose. 
This suggests that the approaching pose given to the trajectory optimizer should be closer to the object.

\captionsetup[subfigure]{labelformat=empty}
\begin{figure*}[]
	\centering
	\subfloat[(a)]{\includegraphics[width=3.4cm]{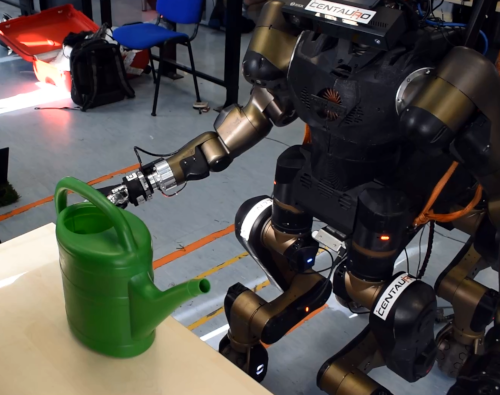}}\hfill
	\subfloat[(b)]{\includegraphics[width=3.4cm]{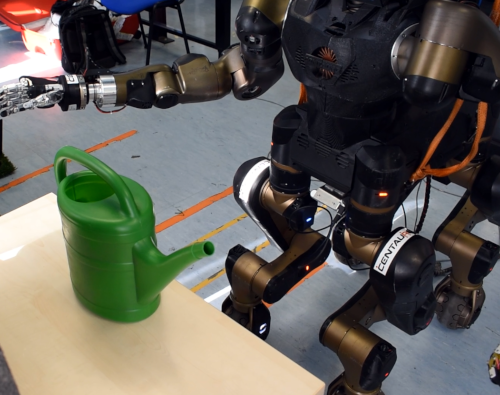}}\hfill
	\subfloat[(c)]{\includegraphics[width=3.4cm]{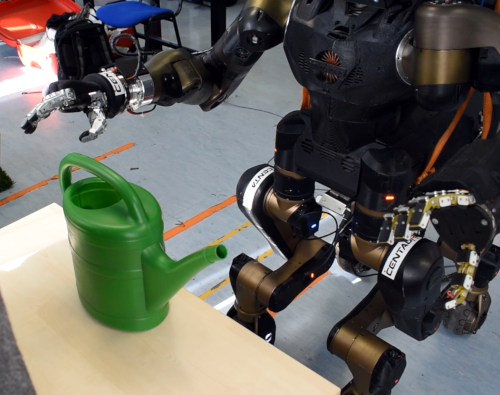}}\hfill
	\subfloat[(d)]{\includegraphics[width=3.4cm]{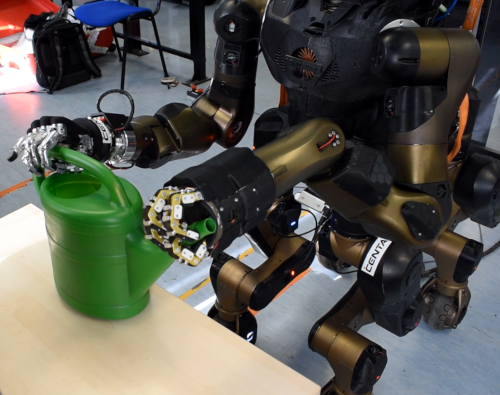}}\hfill
	\subfloat[(e)]{\includegraphics[width=3.4cm]{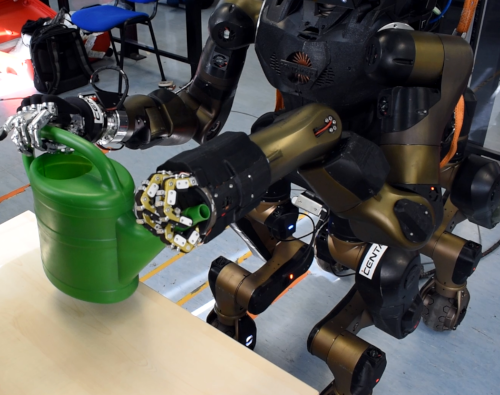}}
	\\
	\subfloat[(a)]{\includegraphics[width=3.4cm]{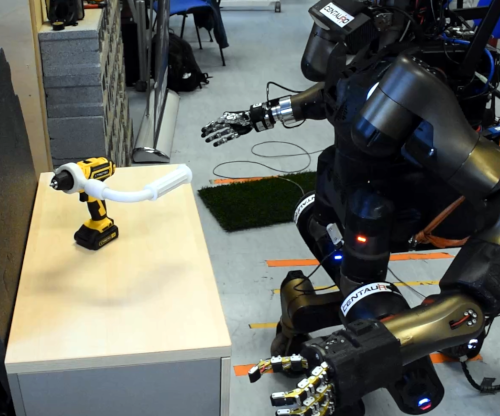}}\hfill
	\subfloat[(b)]{\includegraphics[width=3.4cm]{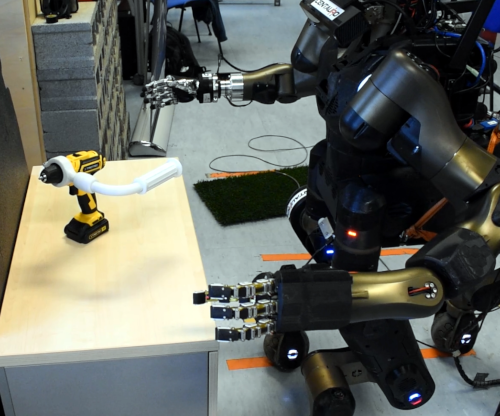}}\hfill
	\subfloat[(c)]{\includegraphics[width=3.4cm]{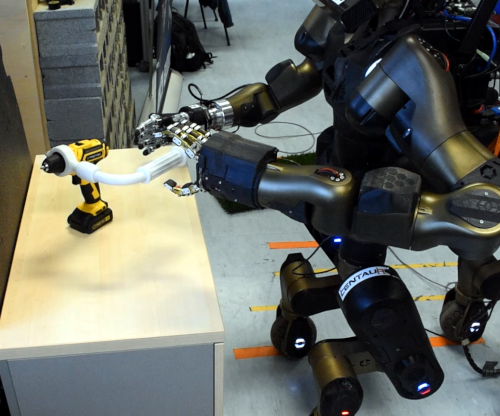}}\hfill
	\subfloat[(d)]{\includegraphics[width=3.4cm]{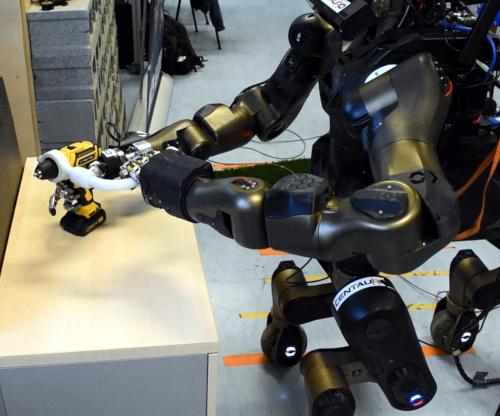}}\hfill
	\subfloat[(e)]{\includegraphics[width=3.4cm]{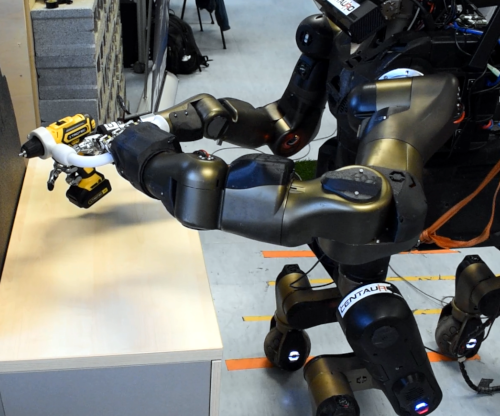}}
	\caption{Centauro performing a dual-arm functional grasp of the watering can and a two-handed drill. \textbf{(a)} Initial pose. \textbf{(b)} - \textbf{(c)} Reaching the pre-grasp pose. 
		\textbf{(d)} Can/drill is grasped. \textbf{(e)} Can/drill is lifted.}
	\label{fig:real_robot_lifting}
\end{figure*}

In addition to the watering can, the Centauro robot also grasped a two-handed drill to demonstrate that our pipeline can be applied to different types of objects.
The process of grasping and lifting both tools is shown in Fig~\ref{fig:real_robot_lifting}.
Footages of the experiments can be found online\footnote{Experiment video: \url{http://www.ais.uni-bonn.de/videos/Humanoids_2018_Bimanual_Manipulation}}.

%% file: conclusions.tex
\section{Conclusions}
We have developed an integrated approach for autonomous dual-arm pick tasks of unknown objects of a know category.
The manipulation pipeline starts with the perception modules, which are capable of segmenting the object of interest.
Given the segmented mesh, we utilize a non-rigid registration method in order to transfer grasps within an object category to the observed novel instance.
Finally, we extended our previous work on STOMP in order to optimize dual-arm trajectories with kinematic chain closure constraint.

We performed a set of experiments in simulation and with the real robot to evaluate the integrated system.
The experiment on trajectory optimization showed that our method can solve the tasks of planning for two arms reliably and fast.
However, with introduction of the closure constraint, the runtime grew significantly.
Nevertheless, we demonstrated that the method is capable of producing feasible trajectories even under multiple complex constraints.
In the simulation experiment, the robot successfully grasped three previously unseen watering cans with two hands from different poses.

On real-robot experiments, our pipeline successfully grasped and lifted several times a watering can and a two-handed drill.
These experiments demonstrated that our system can be successfully applied to solve tasks in the real world in an on-line fashion.